# Learning Norms via Natural Language Teachings

**Taylor Olson**                                    TAYLOROLSON@U.NORTHWESTERN.EDU
**Kenneth D. Forbus**                               FORBUS@NORTHWESTERN.EDU
Department of Computer Science, Northwestern University, Evanston, IL 60208 USA

## Abstract

To interact with humans, artificial intelligence (AI) systems must understand our social world. Within this world norms play an important role in motivating and guiding agents. However, very few computational theories for learning social norms have been proposed. There also exists a long history of debate on the distinction between what is normal (*is*) and what is normative (*ought*). Many have argued that being capable of learning both concepts and recognizing the difference is necessary for all social agents. This paper introduces and demonstrates a computational approach to learning norms from natural language text that accounts for both what is normal and what is normative. It provides a foundation for everyday people to train AI systems about social norms.

## 1. Introduction

Our social world is a game. With the goal of maintaining social approval, we conform to what is common and abide by unwritten rules. Being players of this game by nature, we automatically learn these norms through instruction, observation of action-feedback pairs, and trial and error. As artificial intelligence (AI) systems become even more integrated into society, they must be equipped with these same faculties. Systems must have knowledge of our norms to communicate effectively. For example, elder care systems ought to avoid asking about one's husband if they are aware that he has just passed.

Norms are critical for social cohesion as they constrain desires, allowing for cooperation, as well as guide behavior during uncertainty. Furthermore, being capable of learning norms is crucial, for new situations often call for new guidelines. How then, do we provide AI systems with the means to gain such social and moral knowledge? How do we teach them the rules of the game?

In a recent proposal Malle, Bello, and Scheutz (2019) discuss the importance of norm competence for artificial agents and what is required for such a task. As the authors suggest, simply manually encoding knowledge for such a dynamic and contextual phenomenon is not feasible. Systems must be capable of learning as we do, through rich modalities such as natural language. Important computational work has been done for learning norm representations (Sarathy et al. 2017) but it remains open how these can be learned by artificial agents through more natural modalities. Furthermore, systems must be capable of distinguishing between how frequent a behavior is and its normative status. For example, the act of cheating is quite common, but few would argue that it is permissible. To the best of our knowledge, this capability has not been demonstrated by any current approach.

This paper provides a foundation for AI systems to learn norms via natural language (NL) instruction and testimony. We have constructed a novel frame representation inspired by frame semantics (Fillmore, Wooters, and Baker 2001) that represents both the normative status and prevalence of a norm. We demonstrate our working system that utilizes Dempster-Shafer (DS) theory (Shafer 1976) and performs pragmatic inference via narrative function rules (Tomai and Forbus 2009) to learn norms from language.

We start by providing background on norms and DS theory. We then introduce our qualitative norm frame representation and explain how such frames can be extracted from natural language. Then we show how DS theory can be used to combine evidence for norm frames. Next, we introduce a dataset and show that our approach is capable of learning norms from teachings presented in NL. We conclude with related work and future research.

## 2. Background

### 2.1 What is a norm?

The usage of the term *norm* varies between disciplines. Even within disciplines the concept of normativity varies. Some consider a norm as an Is, or a claim of prevalence (e.g., "Children sometimes share with others."). However, many philosophers have taken the concept of a norm as that of an Ought. This traditional view holds a norm to be claim about what should happen, or an ought-rule (Kelsen 1990). For example, "One should share with their peers." As far back as Hume, philosophers have discussed the need to distinguish between an Is and an Ought (Hume and Levine 2005). Cialdini, Reno, and Kallgren (1990) further argue for the need to distinguish between the Is and the Ought because they refer to separate sources of human motivation. To account for this fundamental distinction, recent theories have unified the concept of a norm by defining two norm types: injunctive and descriptive (Aarts and Dijksterhuis 2003; Cialdini, Reno, and Kallgren 1990). Injunctive norms specify what ought to happen, or, in the case of social norms, what people (dis)approve of. On the other hand, descriptive norms specify what does happen or what most others do. Both types of norms influence human behavior.

Norms have also been shown to be conditional. Though we seem to collectively agree upon some universal norms within a society, such as not harming others, the validity of many norms is conditional upon context. For example, the behavior of wearing clothes is made salient in a public environment but not at the beach (and explains why some attest to having nightmares of going to school without pants!). Cialdini, Reno, and Kallgren (1990) found that subjects littered more in an already littered environment than in a clean environment. Aarts and Dijksterhuis (2003) similarly found that the situation of a library, along with the goal of visiting, activated representations of specific behaviors like being silent.

We often gain knowledge of norms through explicit means such as instruction and testimony. Injunctive norms are relayed to children via instruction (e.g., "You should share with others."). Because descriptive norms instead describe what happens, they are often relayed through testimony of one's own experience (e.g., "People often run in the park."). Other means are more implicit and require further analysis and inference. A child can learn the same sharing norm by observing their sister share her toy, relaying a descriptive norm, and then receiving praise from their mother, providing evidence for an injunctive norm. We also learn norms through second-hand experience from stories, movies, and other media. There is a vast typology of sources for learning norms. Furthermore, within each type, evidence is provided by multiple individual sources. To formalize this evidential reasoning process in machines, we therefore need a mathematical theory of evidence. Here we use Dempster-Shafer (DS) theory, which we overview in the next section.

## 2.2 Dempster-Shafer Theory

Dempster-Shafer (DS) theory (Shafer 1976) is often defined as a generalization of the Bayesian theory of subjective probability. We use Dempster-Shafer theory to explicitly represent epistemic states, such as the evidence for a proposition being mixed, and for combining evidence.

### 2.2.1 Frame of Discernment

DS theory considers an exhaustive set called the frame of discernment (FoD), denoted as $\Theta$, of elements that are mutually exclusive. Each element of the FoD can be interpreted as a possible answer to a question.

### 2.2.2 Basic Belief Assignment

Whereas traditional probability theory would assign a probability just to singletons in the set, DS theory assigns probabilities to the powerset of the frame of discernment, denoted as $2^\Theta$. A basic belief assignment (BBA), or *mass assignment*, is a function, denoted as m(A), that maps each subset of $\Theta$ to a real number in [0,1], such that $m(\emptyset) = 0$ and all assignments sum to 1. Elements with non-zero mass are called *focal elements*.

### 2.2.3 Belief and Plausibility Functions

To represent uncertainty, DS theory computes an interval for a given set of hypotheses. This interval represents a range from the direct evidence for the set and its subsets, to the evidence that is not committed to the negation of the set. The lower and upper limit of these intervals are computed by the *belief function* and the *plausibility function*, respectively.

*Belief:* a function that expresses the total belief of a set and all its subsets. Formally, the belief in a set A is defined as: $Bel(A) = \sum_{B|B \subset A} m(B)$.

*Plausibility:* a function that expresses how much evidence does not contradict the set. Formally, the plausibility of a set is the sum of the basic belief assignments of all sets that intersect that set. The plausibility of a set $A$ can also be computed from belief as: $Pl(A) = 1 - Bel(A^C)$ where $A^C$ is the complement of $A$.

*Uncertainty Interval:* the interval $[Bel(A), Pl(A)]$ represents uncertainty in the set $A$ and it narrows as more evidence in favor of $A$ is accumulated. An example mapping from intervals to an intuitive notion of truth for a proposition $A$ would be: [1,1] = true, [0,0] = false, [0,1] = completely ignorant, [Bel(A), 1] = supported, [0, Pl(A)] = refuted, and [Bel(A), Pl(A)] = evidence for and against.

### 2.2.4 Dempster's Rule of Combination

A fusion operator is used to combine multiple sources of evidence to determine the total support for a given hypothesis. Our approach uses Dempster's rule of combination (Dempster 1967) which computes a new belief function that combines two sources of evidence. Formally, Dempster's rule of combination computes the sum of the mass product intersections. This evidence combination is both associative and commutative and thus evidence can be combined in any order and belief combination can be chained. Dempster's rule is defined as: $m_1 \oplus m_2(c) = \sum_{a \cap b = c} m_1(a) m_2(b) / (1 - K)$, where $K$ represents the conflict between the evidence being combined and is computed as: $K = \sum_{a \cap b = \emptyset} m_1(a) m_2(b)$.

DS theory has been used for combining pieces of evidence for sensor fusion (Premaratne et al. 2009) and in various areas of AI research such as learning indirect speech acts (Wen, Siddiqui, and Williams 2020) and similar norm learning work (Sarathy et al. 2017). There are key properties of DS theory that make it powerful for reasoning about norms. The lack of requirement for priors is a most obvious advantage over Bayes, as it's not clear how one would obtain them in this context. The ability to explicitly represent ignorance is another strength of DS theory. False normative belief can have harmful outcomes, so it is important to distinguish between a reliable source and an unreliable one when using their testimony to reason about norms. Unlike classical probability theory, DS theory allows belief to be "unassigned" to any element (or assigned to the entire frame of discernment) which allows the explicit representation of ignorance. Similarly, another benefit lies in the ability to explicitly represent ambiguity by assigning mass to sets of propositions, rather than just singletons. Imagine a child's first visit to Texas Roadhouse. They walk in and see another kid throw their peanut shells on the floor. Surprised, they look to other people expecting them to provide negative feedback. However, no one responds. At this point, the child is uncertain if this event provides evidence that "throwing peanut shells on the floor in Texas Roadhouse" is obligatory or just optional. Within DS theory you can assign mass directly to that ambiguous set, {Obligatory, Optional}, without having to determine how to distribute the mass across the individual propositions. Though this sort of ambiguity does not arise in the work presented here as norms are introduced explicitly, we note its importance for norm learning in general moving forward.

## 3. Learning Norms from Language

The goal of the norm learning approach presented here is then to learn the evaluation and prevalence of a certain behavior given (or not) a certain context, from teachings presented in natural language. More formally, $E(B)|C$ and $P(B)|C$, where $E$ = Evaluation, $P$ = Prevalence, $B$ = Behavior, $C$ = Context (again, this variable can be unknown or universal). We first describe our qualitative representation for norms and the pragmatic reasoning process that extracts norms from natural language. We then show how we represent belief assignments and perform Dempster-Shafer computations to do such learning.

### 3.1 A Qualitative Norm Frame Representation

We represent norms as frames to tie together the actions, contextual preconditions, and corresponding evaluation and prevalence values. Frames were introduced as data structures for structured knowledge representation and reasoning (Minsky 1975). A frame expresses knowledge of a set of concepts connected to one another via relations or slots. Frame-based representations have been used to represent semantics (Fillmore, Wooters, and Baker 2001) and constructs of qualitative process (QP) theory (McFate, Forbus, and Hinrichs 2014). Here, we represent a norm frame as a conceptual knowledge structure that is connected to other concepts in their slots via four relations: the behavior the norm is about, the context in which the norm is valid, the evaluation of the norm, or how permissible the behavior is in the context, and the prevalence of the norm, or how often the behavior is observed in the given context.

In his discussion of the logical form of events Davidson (1967) pointed out that representing events as predicates is insufficient, as there may be arbitrarily many arguments. He argued instead for representing events as entities. We argue that this is also the case for norms, as we rarely know the contextual preconditions and may only know how frequent a behavior is but not its normative status. Thus, norm frames are also Davidsonian. Our representation requires only that a given norm frame's behavior slot and at least one of either evaluation or prevalence is filled. The context slot is not required. A norm frame can thus be thought of as a rule like structure that maps from an action-scenario to a corresponding frequency and/or normative status.

All the concepts within the norm frame slots are grounded in the NextKB knowledge base (Forbus and Hinrichs 2017). Knowledge in NextKB is partitioned into logical environments defined as Cyc-style microtheories. These microtheories are also hierarchical, so each inherits facts from its parents. This knowledge base contains a large ontology of everyday human actions and locations. These concepts serve as type constraints for the behavior and context slots of a norm frame. The grounded concepts for evaluation consist of the modals from the Traditional Threefold Classification (TTC) of Deontic Logic (McNamara 1996). This scheme posits that there are five normative statuses: three first-order, {Obligatory, Optional, Impermissible} and two second-order, Permissible = {Obligatory, Optional} and Omissible = {Optional, Impermissible}. For prevalence, we define the discrete set of frequencies as: {Continuously, Often, Sometimes, Rarely, Never}. The working second-order frequencies are defined as: MoreThanSometimes = {Continuously, Often}, Sometimes, LessThanSometimes = {Rarely, Never}. Given the deontic and prevalence alphabets, denoted as $\mathbb{D}$ and $\mathbb{P}$ respectively, we define a norm frame as follows:

**Definition** (Norm Frame). A norm frame is a rule like frame structure of the form:

```
(isa ?norm Norm)
(context ?norm ?c)
(behavior ?norm ?b)
(evaluation ?norm ?e)
(prevalence ?norm ?p)
```

where ?c and ?b are concepts for locations and actions/states from NextKB, ?e is in $\mathbb{D}$ and ?p in $\mathbb{P}$.

A reified norm frame thus represents a mapping from a context-behavior pair to qualitative values of permissibility and/or prevalence. Given this definition, descriptive and injunctive norms are defined as follows:

**Definition** (Descriptive Norm). A descriptive norm is a norm frame whose prevalence slot is equal to Often or Continuously.

**Definition** (Injunctive Norm). An injunctive norm is a norm frame whose evaluation slot is equal to Obligatory or Impermissible.

A norm frame can thus be thought of as a "potential norm", as only a subset would be considered as norms based on their evaluation and prevalence slots. Our conception of a norm differs from recent computational formalizations. For example, Sarathy (2020), taking inspiration from Malle et al. (2017), take a norm as "*an instruction to (not) perform action A in context C, provided that a sufficient number of individuals in the community (1) indeed follow this instruction and (2) demand of each other to follow the instruction*." We do not make such an optimal world assumption and instead keep the two conditions separate; the evaluation of a norm frame need not align with the prevalence. As stated earlier, most evaluate the act of cheating as impermissible, but cheating still happens quite often. So, under our norm definitions people can demand others to not perform an action (satisfying demand condition 2), being an injunctive norm, and still fail to follow this instruction by frequently performing the behavior (failing to satisfy prevalence condition 1), being a descriptive norm.

Presented in Figure 1 is an example of the formal predicate calculus representation for the norm of eating on the bus (which most would say is omissible and rarely happens). Again, the concepts *EatingEvent* and *Bus-RoadVehicle* tie into a rich ontology containing collections of human actions, locations, times of day, etc. and many relations.

```
(isa norm1 Norm)
(context norm1 Bus-RoadVehicle)
(behavior norm1 EatingEvent)
(evaluation norm1 Omissible)
(prevalence norm1 Rarely)
```

*Figure 1.* Example norm frame for eating on the bus.

## 3.2 Linguistic Expressions of Norms

It is important to distinguish between a norm as a knowledge construct and its linguistic expression. An imperative (a command, deontic declaration, or an evaluative statement) is the linguistic expression of an injunctive norm. "It expresses an act of will, specifically the meaning that the other person is to behave in a certain way" (Kelsen 1990). Contrarily, testimony of one's own experience, often declarative utterances, express descriptive norms. Such statements can be true or false, which must be distinguished from imperatives. The norm expressed by the statement "People never eat on the bus." is a fact that can be empirically measured, while the norm expressed by "You should not eat on the bus." is not. Statements of testimony like the former express a descriptive norm and the latter an injunctive norm. This distinction also portrays how the concepts of Is and Ought are expressed differently in language. Furthermore, both types of speech acts are merely expressions of, and should be distinguished from, the underlying norm.

## 3.3 Narrative Function

Imperatives and testimony as described here have a typical syntactic and semantic structure. First, they make an evaluative claim (e.g., "You ought to") or provide testimony for a frequency (e.g.,

"People often"). Secondly, they introduce a behavior i.e., the object of a norm. And optionally, they introduce contextual features such as location, time of day, etc. Therefore, when we hear "you should", we then expect the speaker to introduce the behavior being suggested. These expectations help us construct a meaningful and cohesive interpretation. Narrative functions (Barthes 1977; Labov and Waletzky 1996) serve as a representation for such pragmatic constraints that guide understanding. They are acts of a narrator/speaker during dialogue. For example, explaining the setting and introducing a character are both narrative functions. To infer a narrative function is then to infer what information is relevant or what the intended meaning is with respect to context and ongoing discourse.

Detecting narrative functions via abductive rules has been used for understanding natural language sources such as moral decision-making stories (Tomai and Forbus 2009) and extracting QP frame information (McFate, Forbus, and Hinrichs 2014). The narrative functions they compute generate expectations for the system, deriving the intended meaning of the utterances from the lexical, syntactic, and sematic representations. Next, we describe how our model detects specific narrative functions from NL to construct norm frames.

## 3.4 From Natural Language to Norm Frames via Abduction

The process for converting a parse into a norm frame is three-fold: First, the system performs semantic parsing. Second, it extracts norm frame features via abductive rules. Third, norm frames are constructed from the extracted features. We go over each step in the following sections and then provide examples of rules that compute a narrative function.

### 3.4.1 Semantic Parsing, Feature Extraction, and Norm Frame Construction

Our system is built within the Companion cognitive architecture (Forbus & Hinrichs, 2017) which uses the semantic parser CNLU (Tomai and Forbus 2009). CNLU uses Allen's (1994) bottom-up chart parser plus a broad lexicon to create parse trees. It maps from English words to concepts in NextKB (Forbus and Hinrichs 2017) and builds a semantic interpretation of the input using frame semantics extended from FrameNet (Fillmore, Wooters, and Baker 2001). Discourse Representation Theory (Kamp & Reyle, 1993) is used to handle contexts needed for modals and counterfactuals, as well as logical and numerical quantification.

CNLU represents ambiguity in parses in the form of choice sets. A choice set is a disjunctive set of choices for the meaning of terms. Narrative functions are effectively pragmatic rules that constrain these sets, as expectations rule out potential choices. For example, in the sentence "You can run in the park" the word "run" has multiple potential meanings. It could mean fluid flow, as in "The water in the sink was running.", or the human act of running. Clearly, the intended meaning here is the latter. By representing the expectation for a human behavior within imperatives like this, we constrain the set of possible semantics. The choice for fluid flow would be ruled out and the act of running would come through.

After parsing a sentence, our system attempts to abductively prove narrative functions. By proving these statements, the system extracts relevant features and builds corresponding norm frames. We have developed two narrative functions, *IntroductionOfInjunctiveNormEvent* and *IntroductionOfDescriptiveNormEvent*. We define the former as narrative events that introduce an evaluation of a situation or behavior and the latter as events that portray how prevalent a certain situation or behavior is. The reasoner attempts to prove these narrative functions by assuming them, which queries the respective Horn clause rules that analyze the semantics and returns variable bindings. Figure 2 provides an example of converting a semantic parse to a norm frame. This process can be thought of as finding a mapping between a sentence and corresponding concepts that fill the norm frame slots. For example, the narrative function *IntroductionOfInjunctiveNormEvent* is true when there exists a mapping $S \rightarrow (E, B, C)$, where $S$ is the sentence parsed, $E$ is the deontic status mapped from a modal found in the semantics, and $B$ and $C$ are the extracted concepts for behavior and context. Each slot, the evaluation, prevalence, behavior, and context, have their own respective set of detection rules that search for relevant semantic patterns. Because context is optional, if the context detection rules fail, then $C$ is bound to the collection *Location-Underspecified*, the most general concept for locations. Once all extraction rules succeed, another rule constructs a new norm frame, its slot relations, and fills them with the extracted concepts. This frame is then stored in the current discourse microtheory.

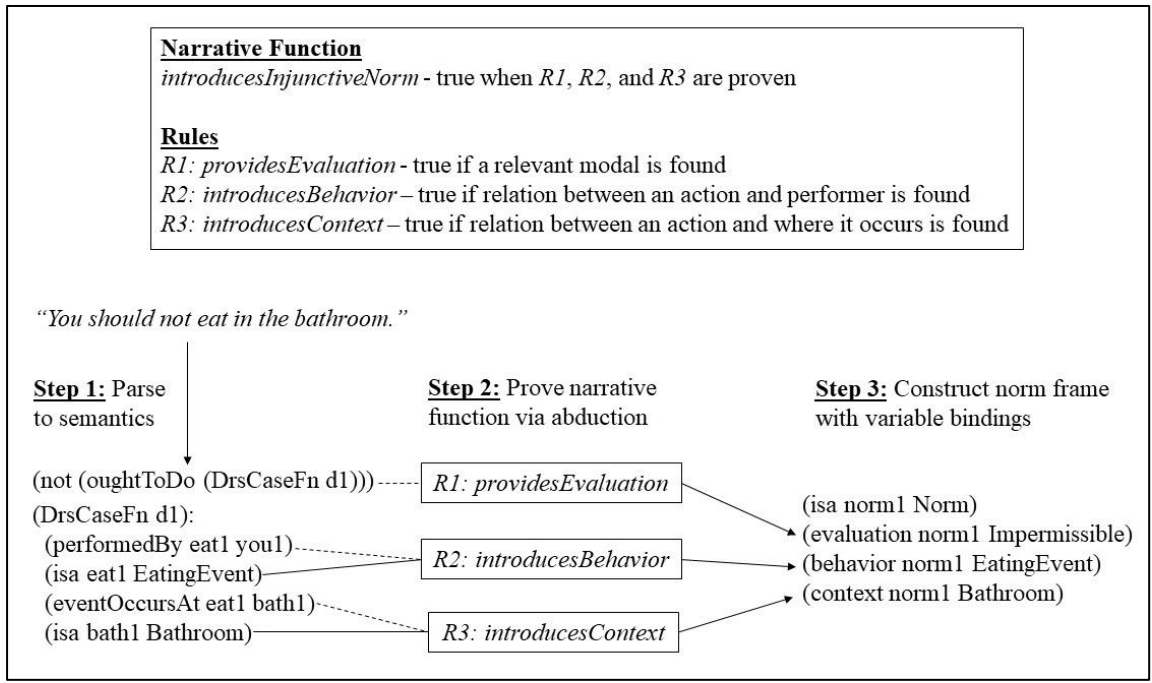

*Figure 2.* Example of constructing a norm frame from a semantic parse.

### 3.4.2  Example Rules

Table 1 provides example rules that detect when a descriptive norm has been introduced. The first rule constructs the logical form for the narrative event and the norm frame from bindings returned by the extraction rules. Narrative events are represented as nonatomic terms like so, *(PresentationEventFn ?sid ?event-id)*. The variable *?sid* denotes the sentence that was processed and *?event-id* is a unique symbol identifier created by the presentation event. The rules for the first antecedent in the second rule, *providesPrevalence*, analyze the syntactic and semantic information produced by the parser CNLU to identify statements of frequency (e.g., "People commonly do X"). As their name suggests, the rules for the second and third antecedents, *introducesBehavior* and *introducesConditionalContext*, detect introductions of behaviors and locations. For example, the most common pattern for behaviors found after parsing is the relation *(performedBy <action> <agent>)* and *(eventOccursAt <action> <location>)* for context. In the rule provided at the bottom of Table 1, the statement *(ist-Information <context> <fact>)* indicates that *<fact>* holds in microtheory *<context>*. Abductive reasoning happens at the level of these *ist-Information* antecedents. The reasoner assumes these statements to be proved, which generates variable bindings.

As stated previously, to handle ambiguity the extraction rules narrow the space of choice sets via type constraints. The rules ensure that the extracted behavior is a specialization of type *Action* or *ConfigurationOfAgent*. The context is ensured to be a specialization of the collection *Location-Underspecified*, indicative of common locations like *LibrarySpace* and *DinnerParty*. These collections tie into a rich ontology that supports reasoning about actions, states, and locations.

*Table 1.* Example rules that compute narrative functions for descriptive norms.

```
(<== (introducesDescriptiveNorm
      (PresentationEventFn ?sid ?event-id)
       ?norm)
  (normMentionedViaPrevalenceStatement
    ?sid ?context ?behavior ?prev)
  (buildNormPrevalenceOnly
    ?sid ?event-id ?context ?behavior ?prev
    ?norm
    (normMentionedViaPrevalenceStatement
      ?sid ?context ?behavior ?prev)))

(<== (normMentionedViaPrevalenceStatement
    ?sid ?context ?behavior ?prev)
  (providesPrevalence ?sid ?prev)
  (introducesBehavior ?sid ?behavior
    ?behavior-var)
  (introducesConditionalContext
    ?sid ?context ?behavior-var))

(<==(providesPrevalence ?sid ?prev)
   (ist-Information (DrsCaseFn ?sid)
     (frequencyOfEventType ?event ?prev)))
```

Table 2 shows the various mappings that result from the narrative function rules for injunctive norms. It is important to note the theoretical difference between causal necessity (a must) and normative necessity (an ought). "What Must I do to realize an end?" is not equivalent to "What Ought I do?" Or as Kelsen (1990) states, "administering poison being a means to killing, does not mean giving poison is an ought". Despite this conceptual distinction, in language we interchangeably use the word "must" for both causal and normative necessity. Similarly, the term "can" (physical possibility) overlaps with "may" (normative possibility).

*Table 2.* Mappings that result from running narrative function rules for injunctive norms.

| Natural Language Input | Deontic Modal |
|---|---|
| Ought to, Should, Must, *Lack of explicit evaluation* (e.g., "Sit in church.") | Obligatory |
| Can, May | Optional |
| Should not, Must not, Cannot, Do not | Impermissible |

These mappings define a heuristic for determining where a given modal lies along the scale of deontic operators. Choosing the weaker element from the scale of deontic elements implies that the speaker believes none of the stronger elements. For example, we argue that within an imperative the terms "can" and "may" do not stand for permissible (in the traditional Deontic sense) but rather optional. When someone states, "You can eat on the bus." they really mean it is optional, not obligatory nor impermissible. Assuming a sort of Gricean (1975) principle, it is the responsibility of the speaker to be as specific as possible, especially when placing themselves in an authoritative role by providing instruction. However, when formed as a question ("Can I eat here?"), the terms do seem to stand for permissibility. In this case, the inquisitor has determined the behavior they wish to perform and are now asking if it is not impermissible i.e., permissible. It is irrelevant if it is something they ought to do or just optional. This capability is necessary for creating systems that can interpret such modal language. A similar discussion can be found in works on scalar/quantity implicatures (Levinson 1983; Carston 1998) and Horn scales (Horn 1972).

### 3.5 Belief-Theoretic Norm Frames

Once a norm frame is extracted from a sentence it is merged with existing evidence. Our framework represents basic belief assignments by explicitly representing evidence for the predicate calculus statements of evaluation and prevalence of a given norm frame. Here, an evidence source is a presentation event resulting from a narrative function. By tracking the collection of evidence for norm frame slots, norms become belief-theoretic, which we define as:

**Definition** (Belief-Theoretic Norm). A belief-theoretic norm is a norm frame with an evaluation and/or prevalence frame of discernment and corresponding body of evidence.

**Definition** (Evaluation Frame of Discernment). The evaluation frame of discernment is a set of possible logical statements of the form:

```
((evaluation ?norm ?val-1) … (evaluation ?norm ?val-n))
```

where *?norm* is a reified norm frame, and *?val-1* to *?val-n* are all the elements of $\mathbb{D}$. The prevalence FoD is analogous, instead with the prevalence predicate and *?val-1* to *?val-n* being all the elements of $\mathbb{P}$.

**Definition** (Body of Evidence). The body of evidence for evaluation and prevalence are sets of mass assignments which take the logical form:

```
(evidenceFor ?pe ?ep ?mass)
```

where *?pe* is a nonatomic term representing a narrative event, *?ep* is a set of statements from the powerset of the evaluation or prevalence FoD, and *?mass* is a mass assignment in the interval [0,1].

Each reified norm introduction event (i.e., the sentence, semantics and interpretation, and meta-information such as the speaker) thus provides a basic belief assignment for the evaluation or prevalence statement(s). Intuitively, it provides evidence that a given norm frame's evaluation and/or prevalence is equal to *?ep*. Because *?ep* is always a singleton, due to lack of ambiguity in direct testimony and instruction, here Dempster-Shafer theory does tend towards traditional probability theory. However, we assign a value of 0.9 for each BBA of a singleton and 0.1 for $\Theta$, representing a measure of trust that is never complete. We plan to incorporate other more ambiguous sources of evidence in the future.

Like Sarathy et al.'s (2017) representation, our approach represents and learns belief-theoretic norms. Here, norm frames are learned by tracking basic belief assignments for the evaluation and prevalence slots. The confidence intervals for each evaluation or prevalence from the respective FoDs are computed by gathering all the evidenceFor statements for the corresponding norm frame slot and then chaining Dempster's rule of combination. We have provided a belief-theoretic norm frame representation for the norm of eating on a bus in Figure 3 below. To save space, we have reduced the bodies of evidence to their resulting confidence intervals.

```
(isa norm1 Norm)
(context norm1 Bus-RoadVehicle)
(behavior norm1 EatingEvent)

Evaluation FoD
[0.04,0.043] (evaluation norm1 Obligatory)
[0.47,0.48] (evaluation norm1 Optional)
[0.47,0.48] (evaluation norm1 Impermissible)

Prevalence FoD
[0.0,0.10] (prevalence norm1 Continuously)
[0.0,0.10] (prevalence norm1 Often)
[0.0,0.1] (prevalence norm1 Sometimes)
[0.9,1.0] (prevalence norm1 Rarely)
[0.0,0.1] (prevalence norm1 Never)
```

*Figure 3.* Belief-theoretic norm frame representation for the act of eating on a bus.

### 3.6 Queries for Reasoning and Question Answering

The system computes belief values for a given norm frame slot via back-chaining queries. We have developed three types of such queries for both evaluation and prevalence. We will describe only those for evaluation, but the queries for prevalence follow the same convention. Each of the following relations take an agent's belief microtheory as the first argument which contains the predicate calculus representations of the norm frames and bodies of evidence. Effectively, these relations represent the epistemic states of an agent given available evidence.

*(believesEvaluationOfBehaviorInContext ?mt ?b ?c ?e):* Determines if a norm is believed in microtheory *?mt*, where belief is true when (bel + pl) /2 ≥ belief threshold. We set the default belief threshold to 0.9. If the behavior *?b* or context *?c* is unbound, it binds them to the concepts from slots of the norm frame that is retrieved. Thus, one can query "Where should I not eat at?" and receive bindings for the variable *?c* consisting of contexts like "in the bathroom".

*(confidenceInEvaluationOfBehaviorInContext ?mt ?b ?c ?e ?interval):* Runs Dempster's rule of combination across the basic belief assignments in microtheory *?mt* for a norm frame with behavior *?b*, context *?c*, and evaluation *?e*. It then computes the belief and plausibility functions, and binds *?confidence-interval* to the interval [belief, plausibility].

*(mostBelievedEvaluationOfBehaviorInContext ?mt ?b ?c ?e):* Runs Dempster's rule of combination for a norm frame with behavior *?b* and context *?c* across the deontic frame of discernment and binds *?e* to the deontic modal(s) with the highest belief value.

### 3.7 Extraction and Learning Algorithm

The full norm extraction and learning algorithm works as follows. It starts with a set of sentences and a symbol denoting the discourse microtheory and the microtheory that holds the agent's beliefs. For each sentence, it performs semantic parsing, resulting in predicate calculus statements being inserted into a discourse microtheory, and then runs the narrative function rules that search for norm introductions in the discourse microtheory. This process constructs norm frames (multiple can be introduced due to left over ambiguity) and stores them in the ongoing discourse. Based on the constructed norm frames for the current sentence, the system then runs a set of plans that perform mass assignment and norm frame merging. This process first ensures norm frames were successfully extracted from the current sentence. If so, it queries the agent's belief microtheory to determine if a norm frame already exists with the same context and behavior as one of the extracted norm frames. If one has already been encountered, it simply records the narrative event as evidence for the relevant slot in this frame. Otherwise, it chooses one of the extracted norm frames and records it in the beliefs microtheory. The evidence statement for this frame is stored as well. This learning process results in the agent's belief microtheory being populated with predicate calculus representations of the extracted norm frames and basic belief assignments from the training data. We show this workflow in figure 4 below.

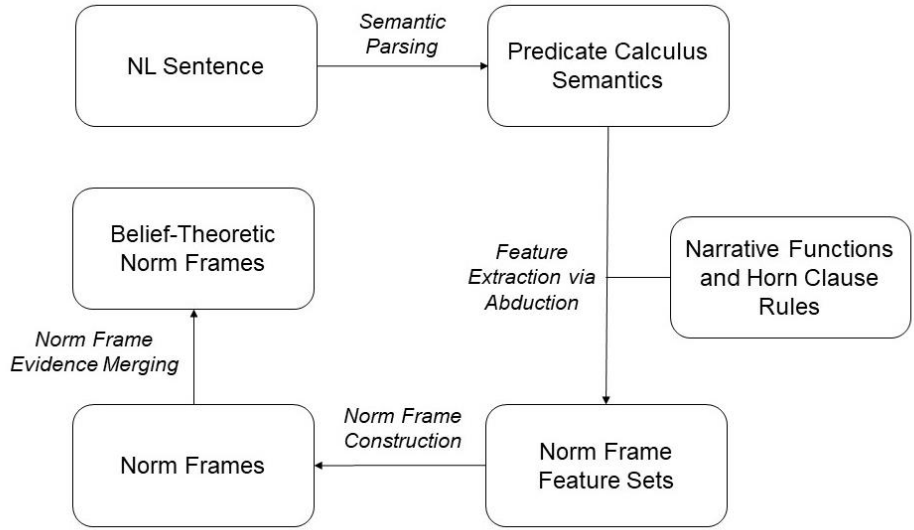

*Figure 4.* Workflow diagram for assimilating evidence from NL teachings.

To further illustrate our norm training framework, let us use the example norm involving eating on the bus. Imagine an agent is told by a bus driver, "You may eat on the bus." The next day the agent is told by a close friend, "You should not eat on the bus." And another friend similarly says, "Do not eat on the bus." Later a stranger advises, "You should eat on the bus." Then while entering the bus a sign reads, "You can eat on the bus." This results in five explicit evidence sources for the same norm frame.

Consider the second friend's instruction, "Do not eat on the bus." After parsing we have the semantics below[1].

```
(not (DrsCaseFn discourse1))
(DrsCaseFn discourse1):
    (performedBy action1 (GapFn :SUBJECT))
    (isa action1 EatingEvent)
    (eventOccursAt action1 context1)
    (isa context1 Bus-RoadVehicle)
```

The negation of a subcontext provides the signal for a negative modal. Querying the narrative functions, a rule will first detect the negative modal and return bindings for the evaluation of Impermissible. The first antecedent for the injunctive norm narrative function rule has been proven, now the rule needs bindings for a behavior and a context. The extraction rules will attempt to abductively prove behavior introductions by assuming: *(ist-Information (DrsCaseFn discourse1) (performedBy ?behavior-var ?agent))* and *(ist-Information (DrsCaseFn discourse1)*

---

[1] Note that the parser does introduce other potential *isa* relations due to ambiguity, but the semantics have been simplified for this example.

*(isa ?behavior-var ?behavior-type))*. This will match the *performedBy* and *isa* relations in the semantics and pass up bindings for the *?behavior-type* as *EatingEvent*, as it satisfies the type constraints i.e., it's a specialization of *Action*. Similarly, tracing through the *eventOccursAt* relation binds the context to *Bus-RoadVehicle*. The system has now proven each antecedent. It has detected a negative modal, a behavior, and a context, and therefore has encountered a negative instruction, or an *IntroductionOfInjunctiveNormEvent*.[2] Now that the narrative function rules have constructed the norm frame, the system searches the agent's belief microtheory and finds that a norm frame, *norm1*, with the same behavior and context has already been learned (from the encounter with the bus driver and the first friend). So, it stores the mass assignment: *(evidenceFor presentation-event1 ((evaluation norm1 Impermissible)) 0.9)*.

Figure 5 shows the confidence in the norm being permissible as the five evidence sources are encountered. As a reminder, DS intervals for permissibility are computed via the back-chaining belief queries containing the equivalent set {Obligatory, Optional}. Figure 5 illustrates how the interval tightens as evidence is encountered. The system has high belief in the fact that eating on a bus is permissible after first talking to the bus driver. However, it is not completely certain yet. As it talks with friends, strangers, and consults other sources it tends to have less belief in the fact that eating on the bus is seen as permissible but becomes more confident in its evaluation. Notice that confidence increases rapidly due to lack of ambiguity in the mass assignments, i.e., a default mass of 0.9 is always assigned to a singleton as previously stated.

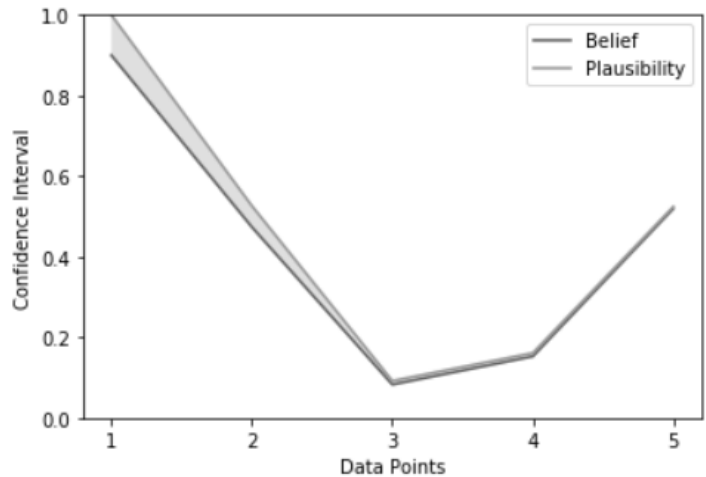

*Figure 5.* Belief and plausibility values overtime for the permissibility of eating on the bus.

---

[2] Some rules require looking within nested discourse structure due to nested modals. For example, processing "You should not X" maps the semantic representation, *(not (oughtToDo X))*, to the deontic modal representation *Impermissible(X)*.

## 4. Evaluation

Evaluating approaches to norm learning is difficult due to the subjective nature of social norms. Therefore, rather than focusing on if the learned norms are desirable themselves, we aimed to evaluate our approach with respect to how well it could extract, track evidence, and compute belief functions for norms presented in natural language.

To test our approach, we curated a training dataset of 100 natural language sentences, some of which we authored, and some were simplified from other sources, such as books on etiquette (Post and Post 2004; Post et al. 2017; Flannery and Sanders 2018) and posts on social norms (Social Norm Examples 2020) and morals (Kittelstad 2020) from the web. We ensured a 50/50 split between imperatives and testimony and that each modal and prevalence pattern was present in the dataset. Two examples from the dataset are, "You should not eat in the bathroom." and "People often sing at recitals." We also constructed a dataset of negative examples to test the closure of our narrative function rules. This dataset consists of 135 sentences that were curated from 4 simplified Wikipedia articles (synopsis of Tell-Tale Heart, description of Mathematics, and biographies of Immanuel Kant and Sojourner Truth) as well as manually constructed atomic stories that aligned with a random subset of the training dataset. For example, we constructed the sentence "Karli yelled in the library" from the deontic declaration "You should not yell in the library." The former sentence does not explicitly introduce a norm, but it does contain the same behavior and context, so it serves as an adversarial example. In summary, the entire training dataset consists of 235 sentences, 100 of which are positive examples (with a 50/50 split between introductions of injunctive and descriptive norms) and 135 being negative examples.

We annotated each positive example with natural language queries and their expected responses. For instance, for the norm of singing, the query and expected response pair generated was: ("How often is someone singing at a recital?", Often). When a certain norm had conflicting sentences endorsing it, we manually labeled the queries with second-order concepts. For example, the set of imperatives contained the sentence, "You should eat in the kitchen.", which provides evidence for obligation. It also contained the sentence, "You can eat in the kitchen.", which endorses the behavior as being optional. In this case we only generate one query for permissibility (obligatory or optional). This means that the number of queries is less than or equal to the amount of datapoints.

To interpret and run the queries, the system parses the sentences into their respective logical forms via the same pragmatic inference process. We have built two simple syntactic rules for detecting norm requests: "What is your evaluation of <behavior, context>?" and "How often is someone <behavior, context>?". To maintain consistency across training and testing, we then use the same behavior and context narrative function extraction rules. To illustrate, "How often is someone singing at a recital?" maps to *(believesPrevalenceOfBehaviorInContext ?beliefs-mt Singing MusicalPerformance ?prev)*.

### 4.1 Experiment One

Our first experiment evaluated the classification accuracy of our narrative function rules. We first ran the norm frame extraction algorithm on the training dataset of natural language sentences and then calculated precision and recall scores. We define a true positive as the event in which a narrative function rule fires on a positive datapoint and a true negative when all rules fail on a negative datapoint.

We received perfect recall over the dataset i.e., norm frames were constructed for each positive example. On the negative examples, the narrative function rules failed as desired on 131 of the 135 sentences, yielding a precision score of 0.96. The four false positives were sentences with common structure to the positive examples. These results yield an F1 score of 0.98.

## 4.2 Experiment Two

In the second experiment, we tested the learning of confidence intervals. Again, we are not concerned with the question of if the agent has desirable beliefs, but rather if our approach yields beliefs that are indicative of the dataset. For example, it should not believe that people never cry at funerals, because there are explicit sentences saying that we often do. Hence, this experiment ensured the norms introduced by each data point were believed by the system after training. To do so, we ran the set of natural language queries with respect to the agent's belief microtheory after training and evaluated for accuracy. If norm extraction, mass assignment, and norm merging all succeed, then each respective belief query should return the correct response. This experiment also evaluates false and relative beliefs due to the nature of DS theory. As a reminder, we determine true belief by: $(bel + pl)/2 \geq 0.9$. By the fact that all masses sum to one, there can be only one element from the frame of discernment with a belief center of mass $\geq 0.9$. Less formally, if the system believes what we say it should, then it also disbelieves what it should.

We achieved 100% query accuracy over the dataset. After training, 47/47 belief queries succeeded for injunctive norms and 47/47 for descriptive norms. Examples of learned norms are provided in table 3 below. Overall, our results suggest that the top-level interpretation rules for modal language are successful at detecting introductions of both norm types. Any failures would result from coverage and ambiguity for behavior and context introductions. These can be handled by extending our language and interpretation coverage and through interactive repair. These results also suggest that by representing norms as frames and storing mass assignments for slots, a system can correctly combine evidence to determine its belief in both norm types. No previous approaches have demonstrated this ability to learn both descriptive and injunctive norms.

*Table 3.* Examples of learned norms.

| Training Sentence(s) | Testing Query | Model Output |
|---|---|---|
| You can eat in the kitchen. You should eat in the kitchen. | What is your evaluation of eating in the kitchen? | Permissible |
| Walk in the hallway. | What is your evaluation of walking in the hallway? | Obligatory |
| You should not steal. | What is your evaluation of someone stealing? | Impermissible |
| People sometimes steal. | How often is someone stealing? | Sometimes |
| People often cry at funerals. | How often is someone crying at funerals? | Often |
| People rarely talk in elevators. | How often is someone talking in elevators? | Rarely |

## 5. Related Work

Our use of narrative function for extraction is analogous to that of McFate, Forbus, and Hinrich's (2014) work on extracting QP frames from natural language. Like them, our largest issue with using rules for narrative functions is broadening coverage. We focused here on single action and location types, but we need to expand to cover mental acts, more sophisticated action and context descriptions, other evaluative statements, and so on. We will continue to extend these extraction rules manually and investigate learning them automatically via ILP techniques and analogy (Crouse, McFate, and Forbus 2018).

Dempster-Shafer theory has also been used for learning, including norms (Sarathy et al. (2017)) and indirect speech acts (Wen, Siddiqui, and Williams's (2020)). However, the data these approaches used came from hand-crafted questionnaires and participant responses. We go beyond these efforts by providing a means to learn from natural language. Furthermore, the evidence representation presented here provides the capability to construct a typology of evidence types to reason over. For instance, a theory may posit that instruction holds more weight for normative reasoning than observation. By reifying the information bearing events in the `evidenceFor` statements, our approach can, for example, determine that a norm is valid due to a single instruction, regardless of how much evidence has been observed via other means.

The most relevant work comes from Sarathy et al.'s (2018) approach to learning cognitive affordances for objects from NL instruction. Narrative function serves a similar function as their pragmatic inference processes. Though their representation of an affordance is like a norm frame in that they are belief-theoretic, norm frames differ in that they contain both deontic status and prevalence as consequents. As discussed throughout this paper, this allows us to represent descriptive and injunctive norms in the same frame, importantly distinguishing between the dimensions of prevalence and evaluation.

## 6. Discussion and Future Work

This paper provides a foundation for teaching artificial agents our norms through natural language instruction. This is important as a step towards ultimately enabling people who are not AI experts to train systems about social norms. We have shown that narrative functions as formal queries can be used to interpret imperatives and testimony presented in natural language. Our work further demonstrates that a combination of DS theory and a norm frame representation allows a system to learn both descriptive and injunctive norms.

We have shown how norms can be learned via explicit means. However, there are many other forms of social learning that introduce norms more implicitly, e.g., observing a mother shake her head shortly after her child yells in the store. Such behavior-response pairs yield evidence for norms as well. We plan to explore utilizing stories as a source of learning norms. The interpretation rules will be built out to perform said story understanding and pragmatic inference. This will also further test the utility of Dempster-Shafer theory as discussed in (Premaratne et al. 2009). We will explore more sophisticated Dempster-Shafer update methods and frameworks (Falkenhainer 1988; Premaratne et al. 2009) and other evidence combination functions (Sentz and Ferson 2002) to support online learning.

Most importantly, we will explore the question of if and how moral norms can be represented and learned by machines. The alert reader might question how this approach (or any norm learning approach) keeps artificial agents from learning undesirable norms (e.g., outdated gender standards, racial stereotypes). As it stands, this approach cannot take such a critical stance on learned norms. To do so, an agent must have a grasp of standards that transcend our cultures and times. We argue that such moral norms must be grounded in more than empirical matters. "That's just not what we do here" is a weak justification for believing that harming others is impermissible. Whether this grounding must come from rational first principles, emotions, or some other source has long been debated (Brennan et al. 2013; Korsgaard 2012; Hume and Levine 2005; Nietzsche 2004). We will analyze such philosophical discussions in hopes to provide a foundation for artificial agents to gain knowledge of moral norms as well.

## Acknowledgements

We thank Constantine Nakos for his helpful feedback. This research was supported by grant FA9550-20-1-0091 from the Air Force Office of Scientific Research.